\documentclass{bmvc2k}

\usepackage{ragged2e}
\usepackage{mathtools}
\usepackage{booktabs}
\usepackage{amsmath}
\usepackage{graphicx}
\usepackage{makecell}
\usepackage{rotating}
\usepackage{multirow}
\usepackage{wrapfig}
\usepackage{lipsum} 
\usepackage{amssymb}
\usepackage{algorithm}
\usepackage{algpseudocode}
\usepackage{amsmath}

\usepackage[export]{adjustbox}

\title{COSMo: \underline{C}LIP Talks on \underline{O}pen-\underline{S}et \underline{M}ulti-Target D\underline{o}main Adaptation}

\addauthor{Munish Monga*}{munish30monga@gmail.com}{1}
\addauthor{Sachin Kumar Giroh*}{22m2159@iitb.ac.in}{1}
\addauthor{Ankit Jha}{ankitjha16@gmail.com}{1,2}
\addauthor{Mainak Singha}{mainaksingha.iitb@gmail.com}{1}
\addauthor{Biplab Banerjee}{getbiplab@gmail.com}{1}
\addauthor{Jocelyn Chanussot}{jocelyn.chanussot@inria.fr}{2}

\addinstitution{
 Indian Institute of Technology, Bombay\\
 Mumbai, India
}
\addinstitution{
 INRIA, Grenoble, France
}

\footnotetext[1]{These authors contributed equally to this work.}

\runninghead{Monga, Giroh, Jha, Singha, Banerjee, Chanussot}{COSMo}

\begin{document}
\maketitle
\vspace{-0.8cm}
\begin{abstract}
Multi-Target Domain Adaptation (MTDA) entails learning domain-invariant information from a single source domain and applying it to multiple unlabeled target domains. Yet, existing MTDA methods predominantly focus on addressing domain shifts within visual features, often overlooking semantic features and struggling to handle unknown classes, resulting in what is known as Open-Set (OS) MTDA. While large-scale vision-language foundation models like CLIP show promise, their potential for MTDA remains largely unexplored. This paper introduces COSMo, a novel method that learns domain-agnostic prompts through source domain-guided prompt learning to tackle the MTDA problem in the prompt space. By leveraging a domain-specific bias network and separate prompts for known and unknown classes, COSMo effectively adapts across domain and class shifts. To the best of our knowledge, COSMo is the first method to address Open-Set Multi-Target DA (OSMTDA), offering a more realistic representation of real-world scenarios and addressing the challenges of both open-set and multi-target DA. COSMo demonstrates an average improvement of $5.1\%$ across three challenging datasets: Mini-DomainNet, Office-31, and Office-Home, compared to other related DA methods adapted to operate within the OSMTDA setting. Code is available at: \href{https://github.com/munish30monga/COSMo}{\texttt{https://github.com/munish30monga/COSMo}}

\end{abstract}

\vspace{-0.8cm}
\section{Introduction}
\label{sec:intro}
\vspace{-0.2 cm}
Domain adaptation (DA) techniques play a crucial role in improving the generalizability of machine learning models across diverse data distributions. DA aims to address the domain shift problem, where models trained on one domain may struggle to generalize effectively to another related domain. Traditionally, DA has mainly focused on a closed-set (CS) setting \cite{csda1, csda2, csda3, csda4}, assuming that the classes in the source and target domains are identical. However, this assumption often does not hold in practical applications, especially in open-world scenarios where class shifts, i.e., unknown classes in the target domain, may occur. 

\begin{figure*}[ht!] 
    \centering
   \includegraphics[width=0.32\textwidth, height = 3.8cm]{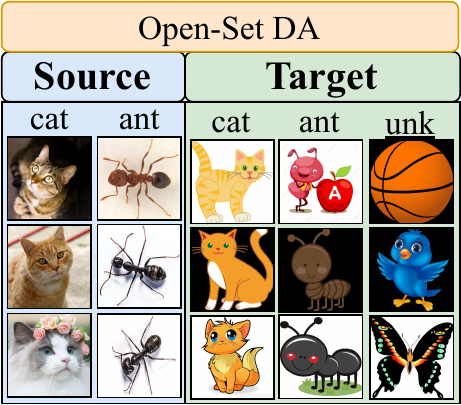}
   \includegraphics[width=0.278\textwidth,height = 3.8cm]{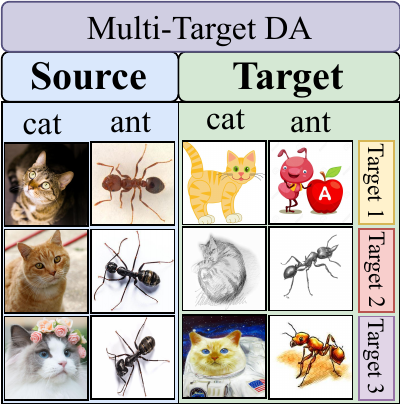}
   \includegraphics[width=0.328\textwidth,height = 3.8cm]{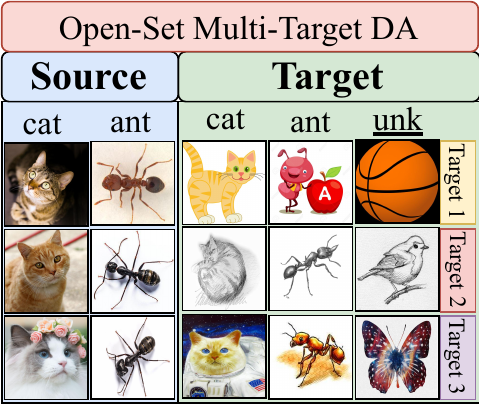}
   \vspace{-0.3cm}
    \caption{OSMTDA differs from traditional DA settings like Open-set DA by handling unknown classes across diverse target domains, while Multi-target DA transfers knowledge from a single labeled source to multiple unlabeled targets. \underline{unk} denotes the unknown class.}
    \vspace{-0.6cm}
    \label{fig:teaser}
\end{figure*}

Open-set DA (OSDA) \cite{OSDA, osda-bp, osda1, osda2, osda3} extends the CS framework by considering classes present in the target domain but absent in the source domain, making it more realistic .

Single-Target DA (STDA) \cite{stda1, stda2, stda3} deals with adapting a model from a single source domain to a single target domain. While effective in controlled settings, STDA encounters challenges in scalability and handling diversity in real-world applications, especially when facing multiple domain shifts simultaneously. Transitioning to a more intricate scenario, Multi-Target DA (MTDA) focuses on adapting a model to perform well across multiple target domains. In situations where domain labels are absent, Blended-Target DA (BTDA) methods \cite{Overwhelm} merge all targets into one for source-target alignment.

In this paper, we introduce Open-set Multi-Target Domain Adaptation (OSMTDA, as shown in Figure \ref{fig:teaser}), merging elements of OSDA and MTDA. OSMTDA tackles three key challenges: managing domain shifts across multiple targets, handling class shifts in open-world scenarios, and adapting from one source to multiple targets. This framework is novel and unexplored. OSMTDA is crucial in domains like autonomous driving, ensuring adaptability across diverse contexts. For instance, vehicles trained in one area may encounter new traffic conditions elsewhere. OSMTDA addresses real-world challenges without extensive labeling or fine-tuning. Additionally, it holds promise in federated learning, bolstering model adaptability while preserving data privacy.

Recently, Vision Language Models (VLMs) \cite{clip,align}, which are trained on a huge amount of data, have shown impressive performance over a wide range of tasks, but they still underperform in downstream tasks. Various techniques are used to make VLMs perform better, such as feature adapter \cite{clip-adapter}, Prompt Learning \cite{cocoop, coop} etc. Textual Prompt Learning in VLMs is effectively utilised in CoOp \cite{coop} where the prompts that are being fed to the text encoder are made learnable, CoCoOp \cite{cocoop} adds on conditional bias, \cite{visualprompt} utilised visual prompts. Maple \cite{maple} utilises both the text and image prompts. Domain Adaptation via Prompt Learning \cite{dapl} utilised prompt learning to solve the domain adaptation task; however, it utilises the domain labels. 

In this paper, we address the OSMTDA problem using source domain-guided prompt learning in VLMs. By combining prompt learning with a domain-specific bias network, we extract knowledge from the source domain.We incorporate domain-specific bias into separate learnable prompts for known and unknown classes to improve the alignment of image and text pairs (discussed in detail in Section \ref{sec:method}). In OSMTDA, we have labeled data from the source domain and unlabeled data from target domains, which include the known classes from the source domain and unlabeled classes as unknown classes. We adopt the similar approach to BTDA \cite{btda}, merging all targets into one and assuming that domain labels are unavailable. We highlight our contributions as:\\
\noindent-To the best of our knowledge, our method is the first to address the task of open-set domain adaptation for multi-target scenarios.\\
\noindent-We propose a source domain-guided prompt learning approach. Separate prompts for known and unknown classes handle the class shift, while a domain-specific bias network addresses the domain shift.\\
\noindent-COSMo outperforms the referred DA methods adapted for OSMTDA, demonstrating significant performance improvement atleast by $5.1\%$ across the Office-Home, Office31, and Mini DomainNet datasets.

\vspace{-0.6cm}
\section{Related Work}
\vspace{-0.2cm}
\label{sec:literature}
\subsection{Open-set Domain Adaptation}
In domain adaptation, the concept of open-set domain adaptation (OSDA) was introduced by \cite{OSDA}. OSDA employs alignment techniques to align the feature spaces of the source and target domains. However, this method relies on unknown source labels, potentially distorting semantic features crucial for class differentiation \cite{disentangled_representations}. In contrast, OSDA-BP \cite{osda-bp} proposes an alternative approach that dispenses with unknown source labels and utilizes adversarial training. This technique enhances the model's ability to distinguish between known and unknown target samples by learning discriminative features invariant across domains. Traditional domain adaptation strategies, such as distribution matching and extracting domain-invariant features, often utilize metrics like Maximum Mean Discrepancy (MMD) to measure domain distances. However, these methods typically overlook the possibility of encountering examples from unknown classes in the target domain, limiting their applicability in open-set scenarios. Additionally, DANCE \cite{dance} presents another notable approach, leveraging neighborhood clustering and entropy-based feature alignment to address the challenges of universal domain adaptation. 

\vspace{-0.4cm}
\subsection{Multi-target domain adaptation}
Multi-Target Domain Adaptation (MTDA) aims to bridge the domain gap by transferring knowledge from a single source domain. This setting has been extensively explored across various tasks, including classification, segmentation \cite{multitarget_seg1,multitarget_seg2}, and object detection. In our research, we adopt a Blended Multi-Target Domain Adaptation approach akin to the framework presented in \cite{Overwhelm}, as we considered the unavailability of domain labels.

Significant efforts in MTDA include using information-theoretic strategies to segregate shared and private information across domains, as implemented in \cite{UMTDA}. Additionally, \cite{UMTDA_segmentation} addresses multi-target domain adaptation for segmentation tasks through a collaborative learning framework. Common strategies for MTDA include adversarial learning, which leverages adversarial networks to minimize domain discrepancies; graph-based methods \cite{curriculum_graph}, which utilize Graph Convolution Networks (GCN) to exploit relational data within and across domains; and knowledge distillation techniques. \textit{In our proposed COSMo, We utilize source domain guided prompt learning to segregate the sets of known and unknown classes in OSMTDA.}

\vspace{-0.4cm}
\subsection{Vision-language models and prompt learning}
Multi-modal vision-language models have made significant strides in various image recognition tasks, utilizing advanced language models like BERT \cite{bert} and GPT \cite{gpt} alongside CNN and ViT for visual analysis. Notable examples include CLIP \cite{clip} and VisualBERT \cite{visualbert}. Traditionally, these models relied on manually crafted textual prompts, which can be complex. Prompt learning methods \cite{coop, cocoop, lasp, maple, gopro} have gained traction for effectively tailoring prompts for downstream tasks. These methods treat token embeddings as learnable variables constrained by image features. Recently, CLIP has been utilized to address challenges in domain adaptation (DA) \cite{adclip} and domain generalization (DG) \cite{stylip, odgclip} tasks. DAPL create disentangled category (class) and domain representations by aligning them differently. \cite{dapl} employs domain-specific context tokens for unsupervised DA, while AD-CLIP \cite{adclip} generates domain-agnostic tokens via a cross-domain style mapping network inspired by \textsc{StyLIP} \cite{stylip}. CLIPN \cite{clipn} tackles out-of-distribution (OOD) tasks by training a "no" text encoder for negative semantic prompts in addition to positive ones. However, these methods are explicitly designed for DG, DA, and OOD tasks. Our model differs from DAPL in the prompt learning technique, also in DAPL, the domain labels are assumed to be known. \textit{In this paper, we utilize CLIP with prompt learning for OSMTDA.}

\vspace{-0.6cm}
\section{Methodology}
\label{sec:method}
\vspace{-0.2cm}
\subsection{Problem formulation}
In the context of OSMTDA, we possess labeled data, denoted as $X_s$, from a single source domain $S$, represented as $(X_s, Y_s) \in S$, and unlabeled data, $X_{t_i}$, from multiple target domains $T_i$, where $X_t = \bigcup\limits_{i=1}^q X_{t_i}$ with $X_{t_i} \in T_i$, and $q$ denotes the number of target domains. Here, $C_s$ and $C_t$ refer to the source and target domain classes, respectively. To establish an open-set scenario, we assume $C_s \subset C_t$, designating the known classes as those from the source domain, denoted as $C_k = C_s$, while the target domain may contain additional unknown classes, denoted as $C_u = C_t \setminus C_s$. For a multi-target setup, we ensure that the target domain classes $C_t$ remain consistent across all $q$ target domains, i.e., $C_t = \{C_{t_i}\}_{i=1}^{q}$. During training, a mini-batch of size $N$ comprises two sets of data: $D_s = {(x_s^i, y_s^i)}_{i=1}^N$ sampled from $(X_s, Y_s)$ and $D_{t} = {(x_{t}^i)}_{i=1}^N$ sampled from $X_{t}$. For a given target image $x_{t}$ from any target domain, the objective in OSMTDA is to accurately classify $x_{t}$ into one of the known classes $C_k$ or identify it as an \texttt{unknown} class.

\vspace{-0.4cm}
\subsection{Overview of COSMo}
\begin{figure*}[ht!]
\begin{center}
  \includegraphics[width=\textwidth]{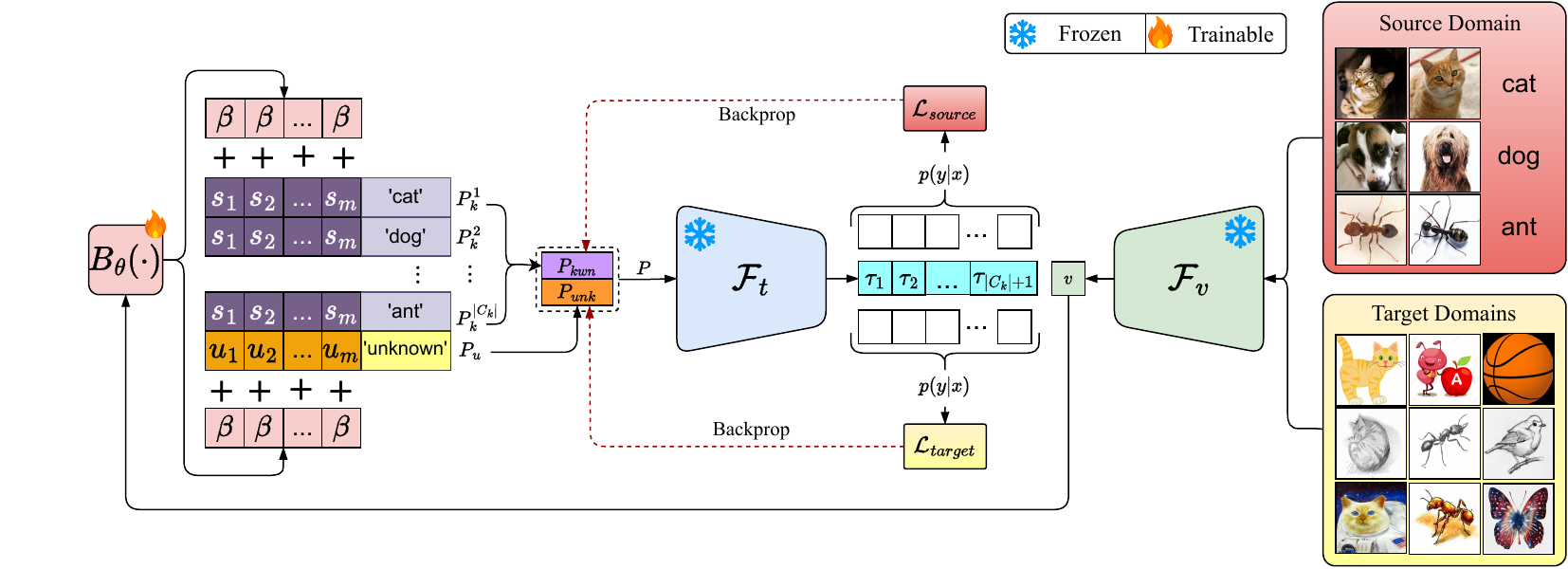} 
\end{center}
\vspace{-0.7cm}
\caption{The architecture overview of COSMo, where $\mathcal{F}_v$ and $\mathcal{F}_t$ are the frozen pretrained CLIP's image and text encoders, respectively. $P_{kwn}$ and $P_{unk}$ denote the prompts for the known and unknown classes, respectively. $\mathcal{B}_\theta(\cdot)$ represents the domain specific bias network, which generates the domain-bias context tokens $\beta$. Best view in color.}
\label{fig:model}
\vspace{-0.5cm}
\end{figure*}

In COSMo, we aim to learn domain-specific and domain-agnostic information. Domain-specific information is learnt via the Domain-Specific Bias Network (DSBN), $(B_\theta(\cdot))$ and the domain-agnostic information is learnt via the separate learnable prompts (known and unknown class-based prompts) for the known and unknown classes. DSBN is trained on both the source and target domain instances, whereas the known and unknown class-based prompts are trained on $D_s$ and $D_t$, respectively, as shown in Figure \ref{fig:model}. We leverage the pre-trained vision encoder $\mathcal{F}_v$ and text encoder $\mathcal{F}_t$ of CLIP. We provide detailed description on the components and working of our proposed COSMo in the following paragraphs.

\noindent\textbf{Domain-Specific Bias Network (DSBN):} DSBN captures domain-specific information from the image features and helps address the domain distribution shift. DSBN is parameterized by \(\theta\), and it modifies the learnable prompts, as the output of \(B_{\theta}\) is directly added to the learnable prompts. The domain information helps in better alignment of the image and text embeddings, as text embedding is based on the unique characteristics of each domain, thus improving the model's adaptability across various domains. \\

\noindent\textbf{Source Domain-Guided Prompt Learning (SDGPL):}  Our approach employs a source domain-guided prompt learning strategy in the prompt space, utilizing a different prompt for the known and unknown classes. The prompts, thus trained, are domain agnostic, and the domain bias is added via the domain-specific bias network.

\begin{itemize}
    \item \textbf{Known class prompts (\(P_{kwn}\))}: Trained on the instances of the source domain classes (\(C_s\)). As depicted in figure \ref{fig:model}, \(P_{k}^c\) captures the domain-agnostic information for class $c$, where \( c \in C_s\) and helps align image-text embedding pairs of the known classes. The known class prompts are constructed as follows:

    \begin{equation}
    P_{k}^c = [s_1][s_2] \ldots [s_m][\text{CLS}]_c, \quad
    P_{k,\ bias}^c = [s_1 + \beta][s_2 + \beta] \ldots [s_m + \beta][\text{CLS}]_c, \\
    \label{eq:prompt_structure}
    \end{equation}
    
    \begin{equation}
    P_{kwn} = [P_{k,\ bias}^1 \ ; P_{k,\ bias}^2 \ ; \ldots \ ; P_{k, \ bias}^{|C_k|}]
    \label{eq:prompt_structure_2}
    \end{equation}
    
    where \(s_i, \text{ for } i \in \{1, \dots, m\}\), represents the \(i^{th}\) context vector (learnable component) of the known class-based prompts and is the same for all the classes in \(C_k\), \( m \) denotes the length of the context prompt, \(\beta\) denotes the domain-bias context token obtained from the domain-specific bias network (\(B_{\theta}\)), \(P_{k,\ bias}^c\) represents the biased known class prompt, \(P_{kwn}\) represents the cumulative prompt for \(|C_{k}|\) classes, and \( [\text{CLS}]_c \) denotes the class vector for class \(c \in C_{k}\).

    \item \textbf{Unknown class prompts (\(P_{unk}\))}: Employed for adapting to and categorizing unknown classes in the target domains. As depicted in figure \ref{fig:model}, unlike \(P_{k}\), \(P_{u}\) is updated through the target domain instances by utilizing the pseudo labels obtained through the gained knowledge on the source domain, thereby enhancing the model's capability to effectively recognize new, unseen categories. The unknown class prompts are constructed as follows:
    
    \begin{equation}
    P_{u} = [u_1][u_2] \ldots [u_{m}][\text{UNK}], \quad P_{unk} = [u_1 + \beta][u_2+ \beta] \ldots [u_{m}+ \beta][\text{UNK}]
    \label{eq:prompt_structure_3}
    \end{equation}
    
    where \( u_i \) represents the context components of the unknown class prompts \(P_{unk}\), \( m \) is the length of the prompt, and \( [\text{UNK}] \) denotes the class vector corresponding to the token "\textit{unknown.}".

\end{itemize} 

\begin{algorithm}
\caption{Pseudo code to train COSMo}
\small
\begin{algorithmic}
\Statex

\Function{model}{$x$}
    \State $v \gets \mathcal{F}_v(x)$ 
    \State $\beta \gets B_\theta(v)$
    \State Get $P_{kwn}$ using $s_i$ and $\beta$, and $P_{unk}$ using $u_i$ and $\beta$
    \Comment{See equations \ref{eq:prompt_structure} and \ref{eq:prompt_structure_3}}
    \State $\tau \gets \mathcal{F}_t([P_{kwn}; P_{unk}])$
    \State $logits \gets \tau_s^T \cdot v$
    \State \Return $logits$
\EndFunction

\Statex
\For{each $(x_s, y_s)$ in $D_s$ and $x_t$ in $D_t$}
    \State $\text{logits}_s \gets \Call{model}{x_s}$
    \Comment{Freeze $u_i$ $\forall$ \( i \in \{1, \dots, m\} \)}
    \State $\mathcal{L}_{\text{source}} \gets \text{Loss}(\text{logits}_s, y_s)$ 
    \Comment{Update $\theta$ and $s_i$ $\forall$ \( i \in \{1, \dots, m\} \)}
    \State Unfreeze $u_i$ $\forall$ \( i \in \{1, \dots, m\} \) \Comment{Unfreeze $u_i$ for target training}
    \State $y_t \gets \text{get\_pseudo\_labels}(x_t)$ \Comment{Get pseudo labels for target data}
    \State $\text{logits}_t \gets \Call{model}{x_t}$
    \Comment{Freeze $s_i$ $\forall$ \( i \in \{1, \dots, m\} \)}
    \State $\mathcal{L}_{\text{target}} \gets \text{Loss}(\text{logits}_t, y_t)$
    \Comment{Update $\theta$ and $u_i$ $\forall$ \( i \in \{1, \dots, m\} \)}
    \State Unfreeze $s_i$ $\forall$ \( i \in \{1, \dots, m\} \) \Comment{Unfreeze $s_i$ for next iteration}
\EndFor
\end{algorithmic}
\label{alg:1}
\end{algorithm}

\vspace{-0.6cm}
\subsection{Model optimization}
In our CLIP-based model, an image \(x\) (from either the source or target domain) is processed by the image encoder $\mathcal{F}_v$, resulting in a feature vector v = $\mathcal{F}_{v}(x)$ and bias (\(\beta =B_{\theta}(\mathcal{F}_{v}(x))\)) is obtained via DSBN. Text prompts with bias ($P_{knw} \ and \ P_{unk}$) are encoded by the text encoder $\mathcal{F}_{txt}$ to get text features \(W\) containing the vectors \(\tau_1, \tau_2, \ldots, \tau_{|C_k|+1}\), where each vector corresponds to the encoded textual representation of a class (including the additional unknown class) with domain bias. The probability of an input image $x$ belonging to class $c$, is computed as:
\begin{equation}
p(y_c|x) = \frac{\exp(\text{sim}(\tau_c,v)/\eta)}{\sum_{i=1}^{|C_k|+1}\exp(\text{sim}(\tau_i, v)/\eta)}
\label{eq1}
\end{equation}
where \(\text{sim}(\tau_c,v)\) represents the similarity between the image feature \(v\) and the text feature for class c is denoted by \(\tau_c\), and \(\eta\) is a temperature parameter that scales the logits before applying the softmax function. The cross-entropy loss for source domain instances, incorporating minimum entropy regularization, is calculated as follows:

\begin{equation}
\mathcal{L}_{\text{source}} = -\mathbb{E}_{(x,y) \sim p(X_S,Y_S)} [\log p(y|x)] +  \lambda \cdot \mathbb{E}_{x \sim p(X_S)} \left[-\sum_{c=1}^{|C_k|+1} p(y_c|x) \log p(y_c|x)\right]
\label{eq2}
\end{equation}

Similarly, we define $\mathcal{L}_{\text{target}}$ as the target loss, which consists of cross entropy and entropy regularization losses, defined in Eq. \ref{eq3}. Here, \(\tilde{y}_c\) represents the obtained pseudo label, \(\lambda\) is a hyperparameter controlling the strength of the entropy regularization. Minimum Entropy regularisation ensures that the model gives prediction with high confidence, whereas the cross entropy loss ensures that the model gives correct prediction.

\begin{equation}
\mathcal{L}_{\text{target}} = -\mathbb{E}_{(x,\tilde{y}) \sim p(X_T,\tilde{Y}_T)} [\log p(y|x)] + \lambda \cdot \mathbb{E}_{x \sim p(X_T)} \left[-\sum_{c=1}^{|C_k|+1} p(\tilde{y}_c|x) \log p(\tilde{y}_c|x)\right]
\label{eq3}
\end{equation}

 
The pseudo-labels are assigned based on the confidence thresholds. More specifically, for an unlabeled instance, if the $p(y_c|x)$ (probability of an input image $x$ belonging to class $c$) for each known class is less than $\kappa_{lower}$, then the instance is labelled as an unknown. Also, if the maximum probability for an unknown class is at least $\kappa_{upper}$, then the instance is labelled as unknown. This selective approach allows the model to focus on more reliable, high-confidence predictions and enhances the overall robustness and accuracy of the domain adaptation process. The training procedure is mentioned in Algorithm \ref{alg:1}. During inference, classification is done with the learnt known and unknown class prompt, assigning the class corresponding to the maximum probability value ($p(y_c|x)$). 

\vspace{-0.4cm}
\section{Experiments}
\label{sec:exper}
\vspace{-0.2cm}
\noindent \textbf{Datasets:} For OSMTDA, we selected three widely used datasets: Office-31 \cite{office31}, Office-Home \cite{officehome}, and Mini-DomainNet \cite{domain_net}. \textbf{Office-31} contains 31 classes across three domains; Amazon (A), DSLR (D), Webcam (W), while \textbf{Office-Home} has 65 classes spanning Art (A), Clip Art (C), Product (P), and Real World (R) domains. \textbf{Mini-DomainNet}, a subset of DomainNet, features 126 classes distributed among Clipart (C), Painting (P), Real (R), and Sketch (S) domains. We ensure an open-set setting by dividing classes into known and unknown categories, maintaining a ratio of $|C_k| / |C_u| = 10/21$ for Office-31, $15/50$ for Office-Home, and $60/66$ for Mini-DomainNet.

\noindent \textbf{Experimental Details and Evaluation Metrics:}
We use the \textit{AdamW} optimizer \cite{kingma2014adam} to optimize COSMo, with a batch size of $32$ and used cosine annealing with an initial learning rate of $0.001$. We utilize two pre-trained vision encoders as $\mathcal{F}_v$: ViT-B/16 \cite{vit} (Table \ref{table:results}) and ResNet-50 \cite{resnet} (results in \texttt{supplementary}). When assessing open-set multi-target domain adaptation, we used commonly employed metrics \cite{HOS, hos2}: average known class accuracy (OS*), the accuracy of unknown classes (UNK), and the harmonic mean score (HOS) between the accuracy of known (OS*) and unknown classes (UNK). In \texttt{supplementary}, we provide the detail explanation of hyperparameter settings of our proposed COSMo.

\vspace{-0.4cm}
\subsection{Comparison to the literature}
To the best of our knowledge, our proposed COSMo is the first attempt at addressing the challenges of OSMTDA, leaving no existing baselines for comparison. Hence for the proposed OSMTDA task, as shown in Table \ref{table:results}, we comprehensively compare COSMo's performance across three distinct datasets against four relevant baseline models: (1) the zero-shot predictions by CLIP \cite{clip} were taken by using standard prompts like: "a dog", "a cat" where class names were provided for the source classes and keeping a $|C_k|$ class classifier, to ensure open-set setting if the maximum probability of predicted class is less than a threshold, it was predicted as unknown. (2) OSDA-BP \cite{OSDA} and (3) DANCE \cite{dance} are used as non-CLIP baselines. Since these methods work under the OSDA setting but with a single target, we merged all target domains into one. (4) AD-CLIP \cite{adclip} also tackles the DA problem with CLIP; but works in a closed-set setup; to ensure open-set setting; we employed the threshold technique as CLIP. COSMo consistently outperforms these methods in terms of the HOS metric, surpassing CLIP, OSDA-BP, DANCE, and AD-CLIP, demonstrating an average improvement of $52.7\%$, $54.6\%$, $5.1\%$, and $83.3\%$, respectively, averaged across all three datasets. We provide detailed results for various domain permutations within each dataset in the \texttt{supplementary} material. \\ \\

\begin{table}[ht!]
\centering
\caption{Comparison of OSMTDA task with our proposed COSMo with state-of-the-art on the Office-31, Office-Home and Mini-DomainNet datasets. The best results are highlighted in \textbf{bold}. We report the HOS metric score. Results are reported using ViT-B/16 backbone.}
\resizebox{\textwidth}{!}{%
\begin{tabular}{@{}lcccc|ccccc|ccccc@{}}
\toprule
\textbf{} & \multicolumn{4}{c}{\textbf{Office-31}} & \multicolumn{5}{c}{\textbf{Office-Home}} & \multicolumn{5}{c}{\textbf{Mini-DomainNet}} \\
\cmidrule(lr){2-5} \cmidrule(lr){6-10} \cmidrule(lr){11-15}
\textbf{Methods} & \textbf{A} & \textbf{D} & \textbf{W} & \textbf{Avg.} & \textbf{A} & \textbf{C} & \textbf{P} & \textbf{R} & \textbf{Avg.} & \textbf{C} & \textbf{P} & \textbf{R} & \textbf{S} & \textbf{Avg.} \\
\midrule
CLIP \cite{clip} &43.79	&39.95&	39.61& 41.12	&61.76&	61.16	&67.24	&63.52	& 63.42&75.78&	75.67&	72.39&	75.66 &74.88 \\
OSDA-BP \cite{osda-bp} &
83.92&	74.60	&73.17&77.23 &	54.39&	33.01&	55.97&	56.95&50.08	&41.27&	52.41&	48.86&	37.79&45.08 \\
DANCE \cite{dance} & 86.48	& 86.34	&87.76	&86.86&\textbf{81.20}	&85.42	&78.58&	79.28& 81.12&	70.26&	81.29&	73.35&	73.84&74.69 \\

AD-CLIP \cite{adclip} &36.97& 48.81& 32.32  &39.37 & 
54.04 & 48.91 & 49.67&46.69 & 49.83&
53.16& 56.92& 47.61& 53.88&52.89  \\
COSMo (Ours) & \textbf{92.46}&	\textbf{88.41}&	\textbf{89.15}&	\textbf{90.01}&80.96&	\textbf{86.8}&	\textbf{82.42}&	\textbf{81.97}&\textbf{83.04} &	\textbf{81.05}&	\textbf{84.15}&	\textbf{79.29}&	\textbf{82.89}& \textbf{81.84}\\
\bottomrule
\end{tabular}
}
\label{table:results}
\end{table}

\begin{figure}[htbp]
    \centering
    \begin{tabular}{ccc}
        \includegraphics[width=0.3\linewidth, height=4cm]{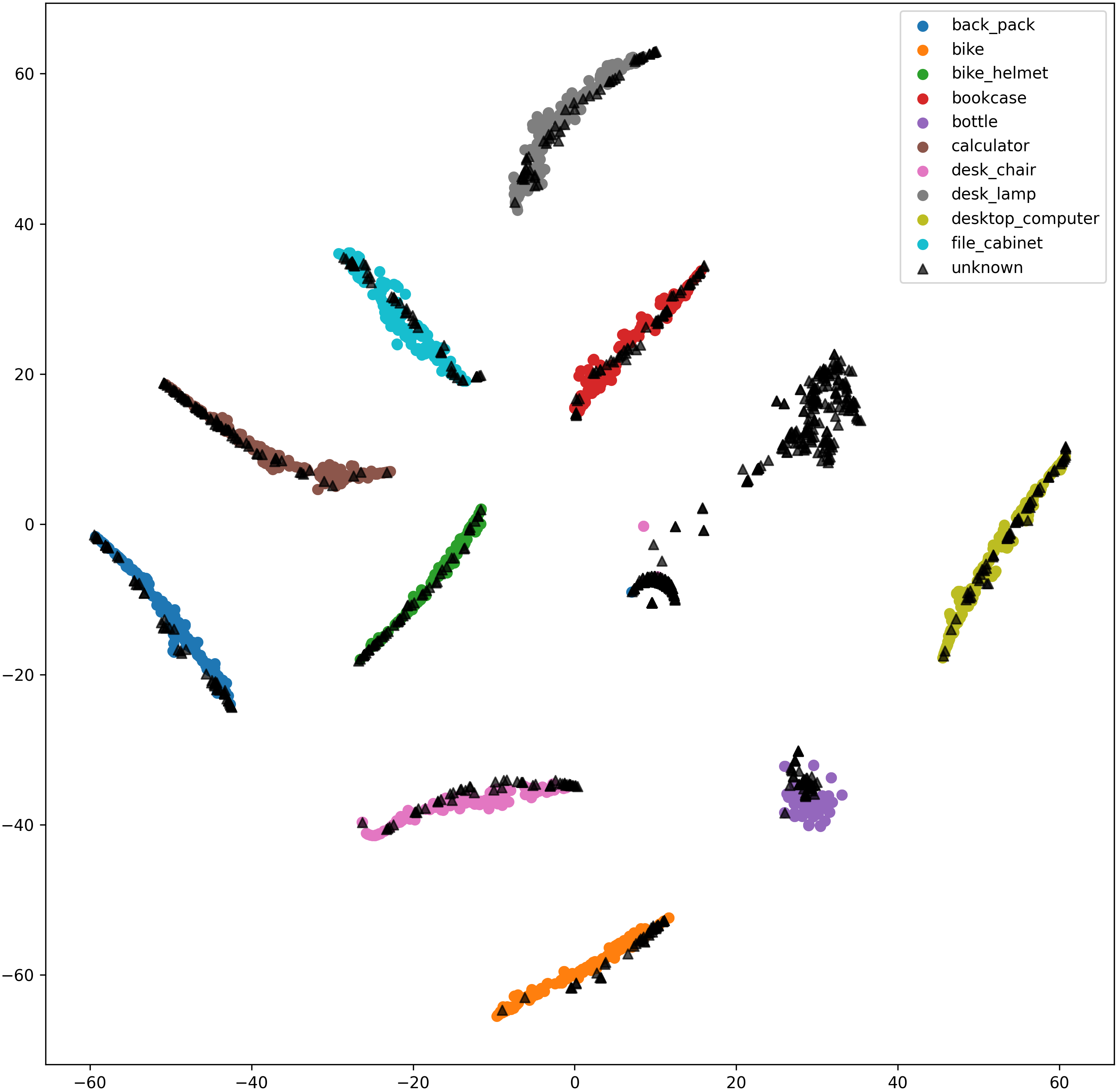} &
        \includegraphics[width=0.3\linewidth, height=4cm]{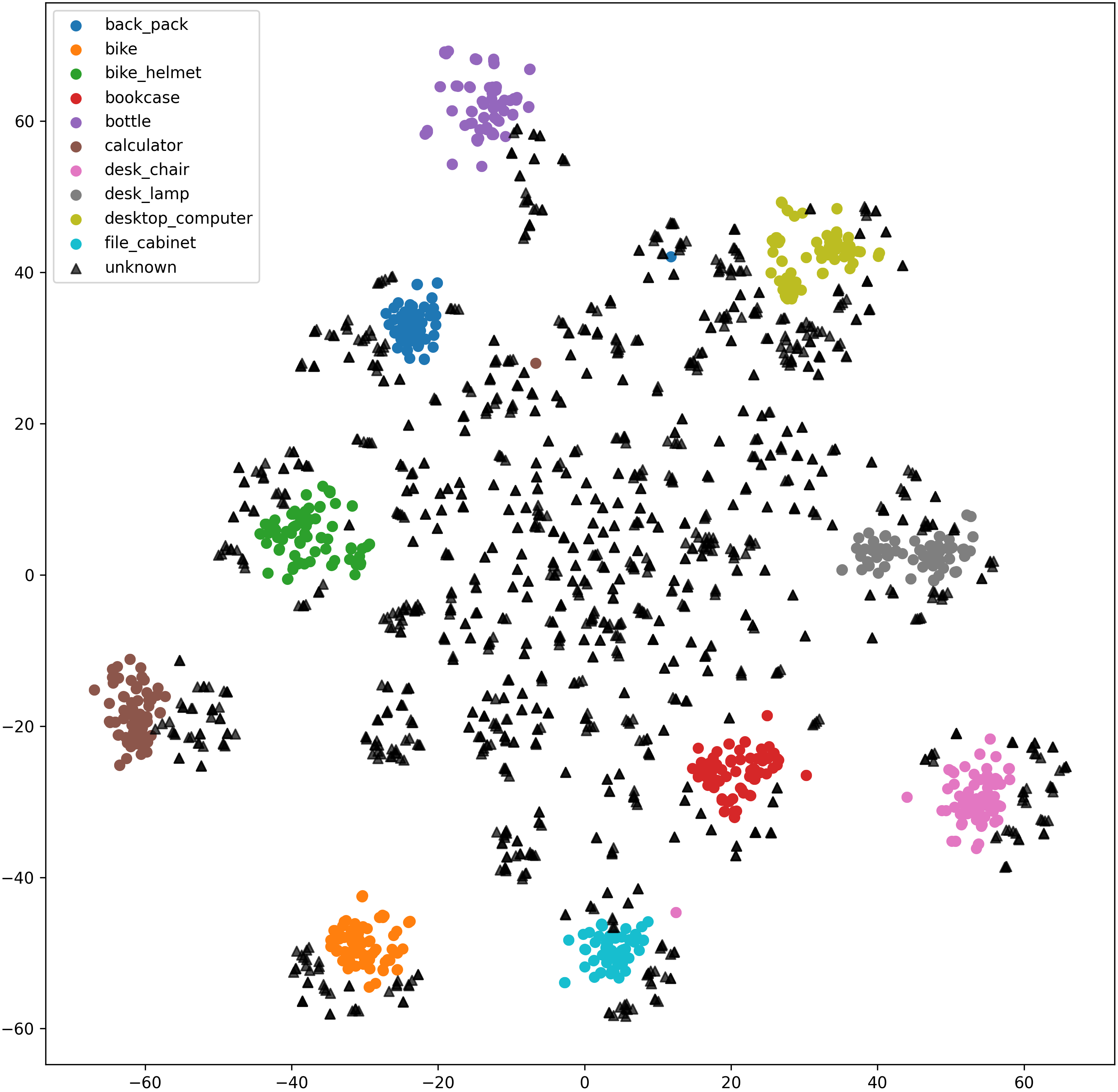} &
        {\includegraphics[width=0.3\linewidth, height=4cm]{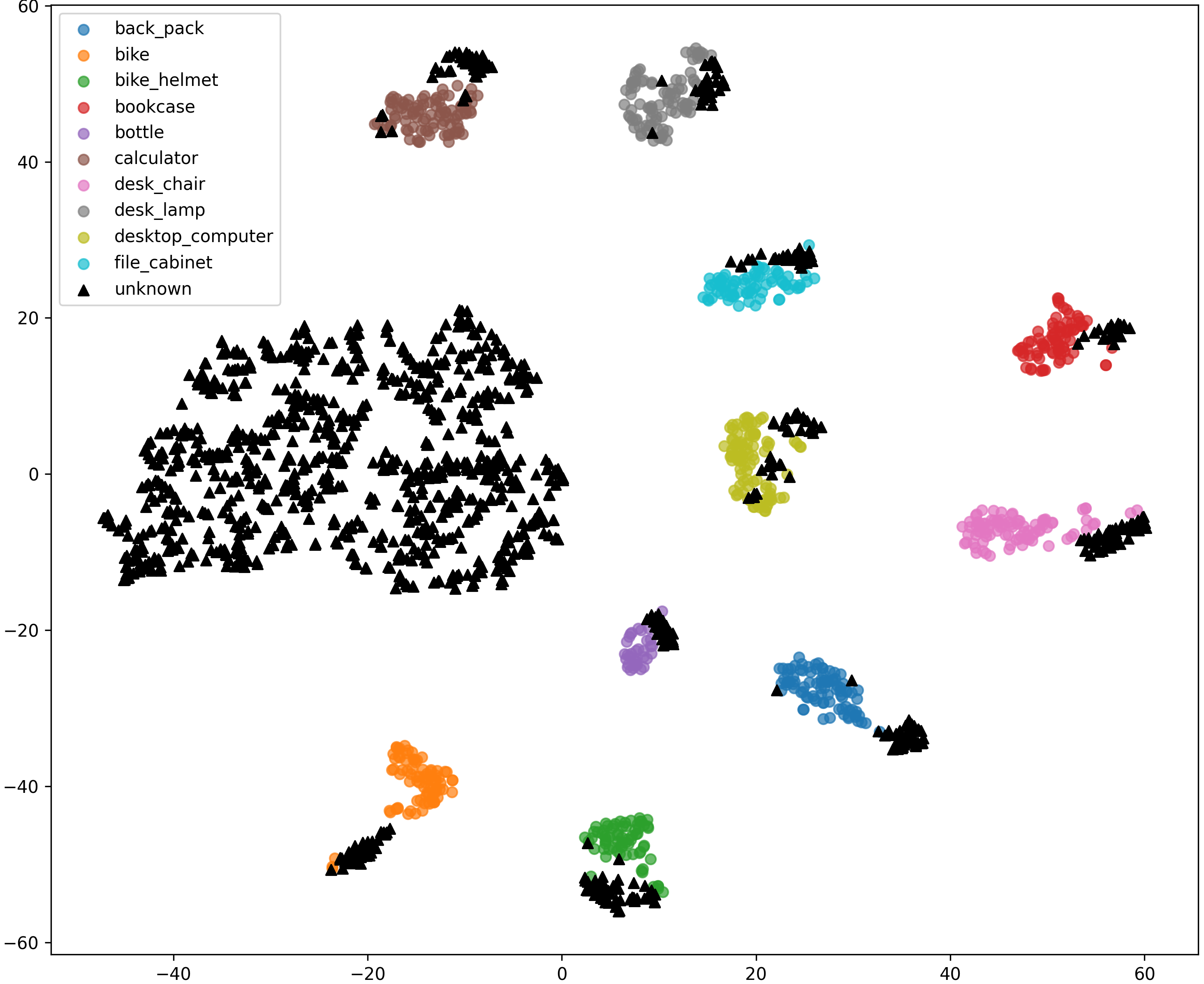}} \\
        
        (a) OSDA-BP & (b) DANCE & (c) COSMo (Ours) \\
    \end{tabular}
    \vspace{-0.2cm}
    \caption{
    t-SNE visualizations on the Office31 Dataset with Amazon as the source domain. Colored dots represent known classes in the source domain, while \textcolor{black}{black} triangles denote target domain samples. For COSMo, text embeddings are used, while features from the penultimate layer are used for the other models.}
    \vspace{-0.6cm}
    \label{fig:tSNEs}
\end{figure}

Furthermore, we visualize the t-SNE embeddings generated from the text encoder of our proposed COSMo for both known and unknown classes, as depicted in Figure \ref{fig:tSNEs}. It is evident that COSMo effectively distinguishes between known and unknown classes across diverse target domains, showcasing superior segregation compared to OSDA-BP \cite{osda-bp} and DANCE \cite{dance}. This observation highlights the robustness of our proposed COSMo in effectively handling both familiar and novel data instances.

\vspace{-0.4cm}
\subsection{Ablation studies}
\noindent\textbf{Impact of having separate $P_{kwn}$ and $P_{unk}$:} Here, we discuss the necessity of having separate prompts for known and unknown classes in our proposed COSMo for the OSMTDA task. From Table \ref{tab:ablation}, it's evident that COSMo with individual prompts for known and unknown classes improves by a margin of $0.38\%$ on the HOS metric. However, experimental findings reveal that using common prompts for known and unknown classes leads to model overfitting to unknown classes, significantly decreasing overall performance. Hence, employing separate prompts plays a crucial role in enhancing the model's ability to effectively adapt to open-set scenarios, thereby improving its overall performance. Further details on the ablations are discussed in the \texttt{supplementary} material. \\
\noindent\textbf{Role of Domain-Specific Bias Network (DSBN):} Additionally, we conduct an ablation study on COSMo, integrating the DSBN network alongside separate prompting for known and unknown classes. Table \ref{tab:ablation} illustrates the impact, showing a significant enhancement in addressing domain shifts for the OSMTDA task. Specifically, the inclusion of DSBN results in an approximate $4\%$ improvement in the HOS metric, underscoring its effectiveness in mitigating domain-related challenges. 

\begin{table}[ht!]
\centering
\caption{Ablations of our proposed COSMo with different components on the Office31 dataset with Amazon as source domain with ViT-B/16 backbone and context length $m$= 4.}
\scalebox{1}{
\begin{tabular}{cccc}
\hline
\textbf{Separate Prompt} & \textbf{DSBN} & \textbf{Entropy Regularisation} & \textbf{HOS} \\ \hline
 \texttimes & \texttimes & \texttimes & 84.41 \\ 
\checkmark & \texttimes & \texttimes & 84.79 \\ 
\checkmark & \checkmark & \texttimes & 88.59 \\ 
\checkmark & \checkmark & \checkmark & \textbf{92.46} \\ \hline
\end{tabular}}
\label{tab:ablation}
\end{table}

\begin{table}[ht!]
\centering
\vspace{-0.2cm}
\caption{Ablations of our proposed COSMo with different context lengths with ViT-B/16 backbone.}
\scalebox{1}{
\begin{tabular}{ccc}
\hline
\textbf{Context length (m)} & \textbf{Trainable Parameters (in K)} & \textbf{HOS} \\ \hline
4 & 37.4 & \textbf{92.46} \\ 
8 & 41.5  & 88.83 \\
16 & 49.7 & 89.58 \\  \hline
\end{tabular}}
\label{tab:ablation_2}
\vspace{-0.6cm}
\end{table}

\noindent \textbf{Ablation with Entropy Regularization loss:}
Finally, we ablate our proposed COSMo model with the loss terms discussed in Eq. \ref{eq2} and \ref{eq3}. Initially, we train the network using only the cross-entropy loss, which fails to classify open-set samples from unseen target domains. Subsequently, we incorporate the entropy regularization loss alongside the cross-entropy loss and train COSMo. Table \ref{tab:ablation} demonstrates that it achieves approximately $4\%$ better HOS performance compared to optimizing the network solely with cross-entropy loss. \\
\noindent\textbf{Ablation with number of context tokens and number of trainable parameters:} We examine the impact of prompt learning in COSMo by varying the context token length with $m = 4, 8$, and $16$. Table \ref{tab:ablation} shows the HOS metric performance on the OSMTDA task, with COSMo trained on the Amazon domain and evaluated on other domains of the Office31 dataset. Notably, $m = 4$ achieves superior performance by a minimum margin of $2.88\%$ across all settings detailed in Table \ref{tab:ablation_2}. Additionally, we report the number of trainable parameters required for training COSMo, observing that with $m=4$, COSMo achieves higher performance with fewer trainable parameters compared to $m=8$ and $16$. 

\vspace{-0.4cm}
\section{Conclusion}
\vspace{-0.2cm}
In this paper, we introduce the novel framework COSMo, addressing the problem of Open-Set Multi-Target Domain Adaptation by leveraging source domain-guided prompt learning. We utilize the frozen image and text encoders of the pre-trained CLIP, along with a few trainable parameters, in designing the network. COSMo incorporates a domain-specific bias network and separate prompts for known and unknown classes, enabling efficient adaptation across domain and class shifts. To our knowledge, we are the first to tackle the challenges of the OSMTDA task, providing a practical representation of real-world scenarios by integrating both open-set and multi-target domain adaptation challenges. While COSMo assumes a consistent set of target domain classes across all domains, real-world scenarios may involve varying sets of classes across target domains. In future work, we aim to extend COSMo to address related tasks such as semantic segmentation and object detection.

\vspace{-0.4cm}
\section{Acknowledgements}
\vspace{-0.2cm}
We gratefully acknowledge the Centre for Machine Intelligence and Data Science (C-MInDS) at the Indian Institute of Technology Bombay for providing the financial support and infrastructure necessary to conduct this research. We also wish to express our appreciation for the valuable feedback provided by the reviewers, which significantly contributed to improving the quality of this paper.

\appendix
\newpage
\bibliography{arxiv}

\begin{thebibliography}{46}
\providecommand{\natexlab}[1]{#1}
\providecommand{\url}[1]{\texttt{#1}}
\expandafter\ifx\csname urlstyle\endcsname\relax
  \providecommand{\doi}[1]{doi: #1}\else
  \providecommand{\doi}{doi: \begingroup \urlstyle{rm}\Url}\fi

\bibitem[Bose et~al.(2024)Bose, Jha, Fini, Singha, Ricci, and Banerjee]{stylip}
Shirsha Bose, Ankit Jha, Enrico Fini, Mainak Singha, Elisa Ricci, and Biplab Banerjee.
\newblock Stylip: Multi-scale style-conditioned prompt learning for clip-based domain generalization.
\newblock In \emph{Proceedings of the IEEE/CVF Winter Conference on Applications of Computer Vision}, pages 5542--5552, 2024.

\bibitem[Bucci et~al.(2020{\natexlab{a}})Bucci, Loghmani, and Tommasi]{HOS}
Silvia Bucci, Mohammad~Reza Loghmani, and Tatiana Tommasi.
\newblock On the effectiveness of image rotation for open set domain adaptation, 2020{\natexlab{a}}.

\bibitem[Bucci et~al.(2020{\natexlab{b}})Bucci, Loghmani, and Tommasi]{osda3}
Silvia Bucci, Mohammad~Reza Loghmani, and Tatiana Tommasi.
\newblock On the effectiveness of image rotation for open set domain adaptation.
\newblock In \emph{European conference on computer vision}, pages 422--438. Springer, 2020{\natexlab{b}}.

\bibitem[Bulat and Tzimiropoulos(2023)]{lasp}
Adrian Bulat and Georgios Tzimiropoulos.
\newblock Lasp: Text-to-text optimization for language-aware soft prompting of vision \& language models.
\newblock In \emph{Proceedings of the IEEE/CVF Conference on Computer Vision and Pattern Recognition}, pages 23232--23241, 2023.

\bibitem[Busto and Gall(2017)]{OSDA}
Pau~Panareda Busto and Juergen Gall.
\newblock Open set domain adaptation.
\newblock In \emph{2017 IEEE International Conference on Computer Vision (ICCV)}, pages 754--763, 2017.
\newblock \doi{10.1109/ICCV.2017.88}.

\bibitem[Cai et~al.(2019)Cai, Li, Wei, Qiao, Zhang, and Hao]{disentangled_representations}
Ruichu Cai, Zijian Li, Pengfei Wei, Jie Qiao, Kun Zhang, and Zhifeng Hao.
\newblock Learning disentangled semantic representation for domain adaptation.
\newblock In \emph{Proceedings of the Twenty-Eighth International Joint Conference on Artificial Intelligence}, IJCAI-2019. International Joint Conferences on Artificial Intelligence Organization, August 2019.
\newblock \doi{10.24963/ijcai.2019/285}.
\newblock URL \url{http://dx.doi.org/10.24963/ijcai.2019/285}.

\bibitem[Chen et~al.(2019)Chen, Zhuang, Liang, and Lin]{btda}
Ziliang Chen, Jingyu Zhuang, Xiaodan Liang, and Liang Lin.
\newblock Blending-target domain adaptation by adversarial meta-adaptation networks, 2019.

\bibitem[Chopra et~al.(2013)Chopra, Balakrishnan, and Gopalan]{csda3}
Sumit Chopra, Suhrid Balakrishnan, and Raghuraman Gopalan.
\newblock Dlid: Deep learning for domain adaptation by interpolating between domains.
\newblock In \emph{ICML workshop on challenges in representation learning}, volume~2. Citeseer, 2013.

\bibitem[Devlin et~al.(2018)Devlin, Chang, Lee, and Toutanova]{bert}
Jacob Devlin, Ming-Wei Chang, Kenton Lee, and Kristina Toutanova.
\newblock Bert: Pre-training of deep bidirectional transformers for language understanding.
\newblock \emph{arXiv preprint arXiv:1810.04805}, 2018.

\bibitem[Dosovitskiy et~al.(2020)Dosovitskiy, Beyer, Kolesnikov, Weissenborn, Zhai, Unterthiner, Dehghani, Minderer, Heigold, Gelly, et~al.]{vit}
Alexey Dosovitskiy, Lucas Beyer, Alexander Kolesnikov, Dirk Weissenborn, Xiaohua Zhai, Thomas Unterthiner, Mostafa Dehghani, Matthias Minderer, Georg Heigold, Sylvain Gelly, et~al.
\newblock An image is worth 16x16 words: Transformers for image recognition at scale.
\newblock \emph{arXiv preprint arXiv:2010.11929}, 2020.

\bibitem[Gao et~al.(2021)Gao, Geng, Zhang, Ma, Fang, Zhang, Li, and Qiao]{clip-adapter}
Peng Gao, Shijie Geng, Renrui Zhang, Teli Ma, Rongyao Fang, Yongfeng Zhang, Hongsheng Li, and Yu~Qiao.
\newblock Clip-adapter: Better vision-language models with feature adapters.
\newblock \emph{arXiv preprint arXiv:2110.04544}, 2021.

\bibitem[Ge et~al.(2022)Ge, Huang, Xie, Lai, Song, Li, and Huang]{dapl}
Chunjiang Ge, Rui Huang, Mixue Xie, Zihang Lai, Shiji Song, Shuang Li, and Gao Huang.
\newblock Domain adaptation via prompt learning.
\newblock \emph{arXiv preprint arXiv:2202.06687}, 2022.

\bibitem[Gholami et~al.(2020)Gholami, Sahu, Rudovic, Bousmalis, and Pavlovic]{UMTDA}
Behnam Gholami, Pritish Sahu, Ognjen Rudovic, Konstantinos Bousmalis, and Vladimir Pavlovic.
\newblock Unsupervised multi-target domain adaptation: An information theoretic approach.
\newblock \emph{IEEE Transactions on Image Processing}, 29:\penalty0 3993--4002, 2020.
\newblock \doi{10.1109/TIP.2019.2963389}.

\bibitem[Gopalan et~al.(2011)Gopalan, Li, and Chellappa]{csda2}
Raghuraman Gopalan, Ruonan Li, and Rama Chellappa.
\newblock Domain adaptation for object recognition: An unsupervised approach.
\newblock In \emph{2011 international conference on computer vision}, pages 999--1006. IEEE, 2011.

\bibitem[He et~al.(2016)He, Zhang, Ren, and Sun]{resnet}
Kaiming He, Xiangyu Zhang, Shaoqing Ren, and Jian Sun.
\newblock Deep residual learning for image recognition.
\newblock In \emph{Proceedings of the IEEE conference on computer vision and pattern recognition}, pages 770--778, 2016.

\bibitem[Hsu et~al.(2015)Hsu, Chen, Hou, Tsai, Yeh, and Wang]{csda4}
Tzu Ming~Harry Hsu, Wei~Yu Chen, Cheng-An Hou, Yao-Hung~Hubert Tsai, Yi-Ren Yeh, and Yu-Chiang~Frank Wang.
\newblock Unsupervised domain adaptation with imbalanced cross-domain data.
\newblock In \emph{Proceedings of the IEEE International Conference on Computer Vision}, pages 4121--4129, 2015.

\bibitem[Isobe et~al.(2021{\natexlab{a}})Isobe, Jia, Chen, He, Shi, Liu, Lu, and Wang]{UMTDA_segmentation}
Takashi Isobe, Xu~Jia, Shuaijun Chen, Jianzhong He, Yongjie Shi, Jianzhuang Liu, Huchuan Lu, and Shengjin Wang.
\newblock Multi-target domain adaptation with collaborative consistency learning, 2021{\natexlab{a}}.

\bibitem[Isobe et~al.(2021{\natexlab{b}})Isobe, Jia, Chen, He, Shi, Liu, Lu, and Wang]{multitarget_seg2}
Takashi Isobe, Xu~Jia, Shuaijun Chen, Jianzhong He, Yongjie Shi, Jianzhuang Liu, Huchuan Lu, and Shengjin Wang.
\newblock Multi-target domain adaptation with collaborative consistency learning, 2021{\natexlab{b}}.

\bibitem[Jang et~al.(2022)Jang, Na, Shin, Ji, Song, and Moon]{hos2}
JoonHo Jang, Byeonghu Na, DongHyeok Shin, Mingi Ji, Kyungwoo Song, and Il-Chul Moon.
\newblock Unknown-aware domain adversarial learning for open-set domain adaptation, 2022.

\bibitem[Jia et~al.(2021)Jia, Yang, Xia, Chen, Parekh, Pham, Le, Sung, Li, and Duerig]{align}
Chao Jia, Yinfei Yang, Ye~Xia, Yi-Ting Chen, Zarana Parekh, Hieu Pham, Quoc Le, Yun-Hsuan Sung, Zhen Li, and Tom Duerig.
\newblock Scaling up visual and vision-language representation learning with noisy text supervision.
\newblock In \emph{International Conference on Machine Learning}, pages 4904--4916. PMLR, 2021.

\bibitem[Khattak et~al.(2023)Khattak, Rasheed, Maaz, Khan, and Khan]{maple}
Muhammad~Uzair Khattak, Hanoona Rasheed, Muhammad Maaz, Salman Khan, and Fahad~Shahbaz Khan.
\newblock Maple: Multi-modal prompt learning.
\newblock In \emph{Proceedings of the IEEE/CVF Conference on Computer Vision and Pattern Recognition (CVPR)}, pages 19113--19122, June 2023.

\bibitem[Kingma and Ba(2014)]{kingma2014adam}
Diederik~P Kingma and Jimmy Ba.
\newblock Adam: A method for stochastic optimization.
\newblock \emph{arXiv preprint arXiv:1412.6980}, 2014.

\bibitem[Kundu et~al.(2020)Kundu, Venkat, Revanur, Babu, et~al.]{osda2}
Jogendra~Nath Kundu, Naveen Venkat, Ambareesh Revanur, R~Venkatesh Babu, et~al.
\newblock Towards inheritable models for open-set domain adaptation.
\newblock In \emph{Proceedings of the IEEE/CVF conference on computer vision and pattern recognition}, pages 12376--12385, 2020.

\bibitem[Li et~al.(2019)Li, Yatskar, Yin, Hsieh, and Chang]{visualbert}
Liunian~Harold Li, Mark Yatskar, Da~Yin, Cho-Jui Hsieh, and Kai-Wei Chang.
\newblock Visualbert: A simple and performant baseline for vision and language.
\newblock \emph{arXiv preprint arXiv:1908.03557}, 2019.

\bibitem[Liu et~al.(2023)Liu, Li, Wu, and Lee]{visualprompt}
Haotian Liu, Chunyuan Li, Qingyang Wu, and Yong~Jae Lee.
\newblock Visual instruction tuning, 2023.

\bibitem[Long et~al.(2016)Long, Zhu, Wang, and Jordan]{stda1}
Mingsheng Long, Han Zhu, Jianmin Wang, and Michael~I Jordan.
\newblock Unsupervised domain adaptation with residual transfer networks.
\newblock \emph{Advances in neural information processing systems}, 29, 2016.

\bibitem[Long et~al.(2017)Long, Zhu, Wang, and Jordan]{stda2}
Mingsheng Long, Han Zhu, Jianmin Wang, and Michael~I Jordan.
\newblock Deep transfer learning with joint adaptation networks.
\newblock In \emph{International conference on machine learning}, pages 2208--2217. PMLR, 2017.

\bibitem[Panareda~Busto and Gall(2017)]{osda1}
Pau Panareda~Busto and Juergen Gall.
\newblock Open set domain adaptation.
\newblock In \emph{Proceedings of the IEEE international conference on computer vision}, pages 754--763, 2017.

\bibitem[Peng et~al.(2019)Peng, Bai, Xia, Huang, Saenko, and Wang]{domain_net}
Xingchao Peng, Qinxun Bai, Xide Xia, Zijun Huang, Kate Saenko, and Bo~Wang.
\newblock Moment matching for multi-source domain adaptation, 2019.

\bibitem[Radford et~al.(2018)Radford, Narasimhan, Salimans, Sutskever, et~al.]{gpt}
Alec Radford, Karthik Narasimhan, Tim Salimans, Ilya Sutskever, et~al.
\newblock Improving language understanding by generative pre-training.
\newblock \emph{https://www.mikecaptain.com/resources/pdf/GPT-1.pdf}, 2018.

\bibitem[Radford et~al.(2021)Radford, Kim, Hallacy, Ramesh, Goh, Agarwal, Sastry, Askell, Mishkin, Clark, et~al.]{clip}
Alec Radford, Jong~Wook Kim, Chris Hallacy, Aditya Ramesh, Gabriel Goh, Sandhini Agarwal, Girish Sastry, Amanda Askell, Pamela Mishkin, Jack Clark, et~al.
\newblock Learning transferable visual models from natural language supervision.
\newblock In \emph{International Conference on Machine Learning}, pages 8748--8763. PMLR, 2021.

\bibitem[Roy et~al.(2021)Roy, Krivosheev, Zhong, Sebe, and Ricci]{curriculum_graph}
Subhankar Roy, Evgeny Krivosheev, Zhun Zhong, Nicu Sebe, and Elisa Ricci.
\newblock Curriculum graph co-teaching for multi-target domain adaptation, 2021.

\bibitem[Saenko et~al.(2010{\natexlab{a}})Saenko, Kulis, Fritz, and Darrell]{csda1}
Kate Saenko, Brian Kulis, Mario Fritz, and Trevor Darrell.
\newblock Adapting visual category models to new domains.
\newblock In \emph{Computer Vision--ECCV 2010: 11th European Conference on Computer Vision, Heraklion, Crete, Greece, September 5-11, 2010, Proceedings, Part IV 11}, pages 213--226. Springer, 2010{\natexlab{a}}.

\bibitem[Saenko et~al.(2010{\natexlab{b}})Saenko, Kulis, Fritz, and Darrell]{office31}
Kate Saenko, Brian Kulis, Mario Fritz, and Trevor Darrell.
\newblock Adapting visual category models to new domains.
\newblock In \emph{European conference on computer vision}, pages 213--226. Springer, 2010{\natexlab{b}}.

\bibitem[Saito et~al.(2018)Saito, Yamamoto, Ushiku, and Harada]{osda-bp}
Kuniaki Saito, Shohei Yamamoto, Yoshitaka Ushiku, and Tatsuya Harada.
\newblock Open set domain adaptation by backpropagation, 2018.

\bibitem[Saito et~al.(2020)Saito, Kim, Sclaroff, and Saenko]{dance}
Kuniaki Saito, Donghyun Kim, Stan Sclaroff, and Kate Saenko.
\newblock Universal domain adaptation through self-supervision.
\newblock 2020.

\bibitem[Saporta et~al.(2021)Saporta, Vu, Cord, and Pérez]{multitarget_seg1}
Antoine Saporta, Tuan-Hung Vu, Matthieu Cord, and Patrick Pérez.
\newblock Multi-target adversarial frameworks for domain adaptation in semantic segmentation, 2021.

\bibitem[Singha et~al.(2023{\natexlab{a}})Singha, Jha, and Banerjee]{gopro}
Mainak Singha, Ankit Jha, and Biplab Banerjee.
\newblock Gopro: Generate and optimize prompts in clip using self-supervised learning.
\newblock \emph{arXiv preprint arXiv:2308.11605}, 2023{\natexlab{a}}.

\bibitem[Singha et~al.(2023{\natexlab{b}})Singha, Pal, Jha, and Banerjee]{adclip}
Mainak Singha, Harsh Pal, Ankit Jha, and Biplab Banerjee.
\newblock Ad-clip: Adapting domains in prompt space using clip.
\newblock In \emph{Proceedings of the IEEE/CVF International Conference on Computer Vision}, pages 4355--4364, 2023{\natexlab{b}}.

\bibitem[Singha et~al.(2024)Singha, Jha, Bose, Nair, Abdar, and Banerjee]{odgclip}
Mainak Singha, Ankit Jha, Shirsha Bose, Ashwin Nair, Moloud Abdar, and Biplab Banerjee.
\newblock Unknown prompt the only lacuna: Unveiling clip's potential for open domain generalization.
\newblock In \emph{Proceedings of the IEEE/CVF Conference on Computer Vision and Pattern Recognition}, pages 13309--13319, 2024.

\bibitem[Venkateswara et~al.(2017)Venkateswara, Eusebio, Chakraborty, and Panchanathan]{officehome}
Hemanth Venkateswara, Jose Eusebio, Shayok Chakraborty, and Sethuraman Panchanathan.
\newblock Deep hashing network for unsupervised domain adaptation.
\newblock In \emph{Proceedings of the IEEE conference on computer vision and pattern recognition}, pages 5018--5027, 2017.

\bibitem[Wang et~al.(2023)Wang, Li, Yao, and Li]{clipn}
Hualiang Wang, Yi~Li, Huifeng Yao, and Xiaomeng Li.
\newblock Clipn for zero-shot ood detection: Teaching clip to say no, 2023.

\bibitem[Xu et~al.(2023)Xu, Wang, and Ling]{Overwhelm}
Pengcheng Xu, Boyu Wang, and Charles Ling.
\newblock Class overwhelms: Mutual conditional blended-target domain adaptation, 2023.

\bibitem[Yao et~al.(2015)Yao, Pan, Ngo, Li, and Mei]{stda3}
Ting Yao, Yingwei Pan, Chong-Wah Ngo, Houqiang Li, and Tao Mei.
\newblock Semi-supervised domain adaptation with subspace learning for visual recognition.
\newblock In \emph{Proceedings of the IEEE conference on Computer Vision and Pattern Recognition}, pages 2142--2150, 2015.

\bibitem[Zhou et~al.(2022{\natexlab{a}})Zhou, Yang, Loy, and Liu]{cocoop}
Kaiyang Zhou, Jingkang Yang, Chen~Change Loy, and Ziwei Liu.
\newblock Conditional prompt learning for vision-language models.
\newblock In \emph{Proceedings of the IEEE/CVF Conference on Computer Vision and Pattern Recognition}, pages 16816--16825, 2022{\natexlab{a}}.

\bibitem[Zhou et~al.(2022{\natexlab{b}})Zhou, Yang, Loy, and Liu]{coop}
Kaiyang Zhou, Jingkang Yang, Chen~Change Loy, and Ziwei Liu.
\newblock Learning to prompt for vision-language models.
\newblock \emph{International Journal of Computer Vision}, 130\penalty0 (9):\penalty0 2337--2348, 2022{\natexlab{b}}.

\end{thebibliography}
\newpage
\section{Contents of the supplementary materials}
\vspace{-0.2cm}
We discuss the following aspects in the supplementary:
\begin{itemize}
    \item We describe the dataset details, split ratio, and dataset statistics for the Open-Set Multi-Target Domain Adaptation (OSMT-DA) in Section \ref{sec:dataset} (Table \ref{table:sample}).
  
    \item In Section \ref{sec:ablation}, we provide the ablation study on the effect of the entropy regularization parameter and separate prompts, as shown in Figures \ref{fig:lambda_effect} and \ref{fig:sep_prompts}, respectively.

\item In Section \ref{sec:results}, we present comprehensive results for the three datasets used in our work. Tables \ref{table:results1}, \ref{table:results2}, and \ref{table:results3} report the performance on the Office-31, Office-Home, and Mini-DomainNet datasets, respectively, using OS, OS*, and UNK as evaluation metrics. We also provide a comparison between the t-SNE visualizations from our proposed method, COSMo, and state-of-the-art methods on the Office-Home dataset, shown in Figure \ref{fig:tSNEs}.

\item Finally, in Table \ref{tab:notations}, we list the notations used in designing and training the architecture.

\end{itemize}
\vspace{-1cm}
\section{Dataset Statistics}
\vspace{-0.2cm}
\label{sec:dataset}
Table~\ref{table:sample} presents the distribution of known and unknown samples across different source domains for three datasets: Office, Office-Home, and Mini-DomainNet. It details the counts of known and unknown samples for each domain within these datasets, providing a clear overview of the data variability and composition used in our analysis. 

To assess the efficacy of our approach in open-set Multi-Target domain adaptation, we utilize three established datasets, each offering distinct challenges and settings. The \textbf{Office-Home} dataset \cite{officehome} consists of 15,500 images across four distinct domains: Art, Clip Art, Product, and Real World. It encompasses 65 categories depicting a variety of objects typically found in office and home environments. The \textbf{Office-31} dataset \cite{office31} includes 4,652 images spanning three domains: Amazon, DSLR, and Webcam, with each domain featuring 31 categories related to office supplies. 
Lastly, the \textbf{Mini-DomainNet}, a subset of the larger DomainNet \cite{domain_net} dataset, provides a broad spectrum of images across four domains—Clipart, Painting, Real, and Sketch—comprising 126 classes. 
Dataset split ($|C_k| / |C_u|$) for Office-31, Office-Home and Mini-DomainNet is taken as $10/21$, $15/50$ and $60/66$ respectively. 

\begin{table}[ht!]
\centering
\vspace{-0.2cm}
\caption{Statistics for each dataset depicting the number of known and unknown samples for each source domain}
\begin{tabular}{cccc}
\toprule
\textbf{Dataset} & \textbf{Source Domain} & \textbf{\# known samples} & \textbf{\# unknown samples} \\
\hline
\multirow{3}{*}{Office-31} & Amazon (A) & 389 & 904 \\
 & DSLR (D) & 1059 & 2553 \\
 & Webcam (W) & 978 & 2337 \\
\hline
\multirow{4}{*}{Office-Home} & Art (A) & 3396 & 9765 \\
 & Clipart (C) & 3023 & 8200 \\
 & Product (P) & 3062 & 8087 \\
 & Real (R) & 2936 & 8295 \\
\hline

\multirow{4}{*}{Mini-DomainNet} & Clipart (C) & 55334 & 71108 \\
 & Painting (P) & 50467 & 63176 \\
 & Real (R) & 30524 & 44263 \\
 & Sketch (S) & 54322 & 66241 \\
\bottomrule
\end{tabular}

\label{table:sample}
\vspace{-0.4cm}
\end{table}

\vspace{-0.4cm}
\section{Ablations}
\label{sec:ablation}
\begin{figure}[ht!]
    \centering
    \vspace{-0.2cm}
    \includegraphics[width=0.7\textwidth]{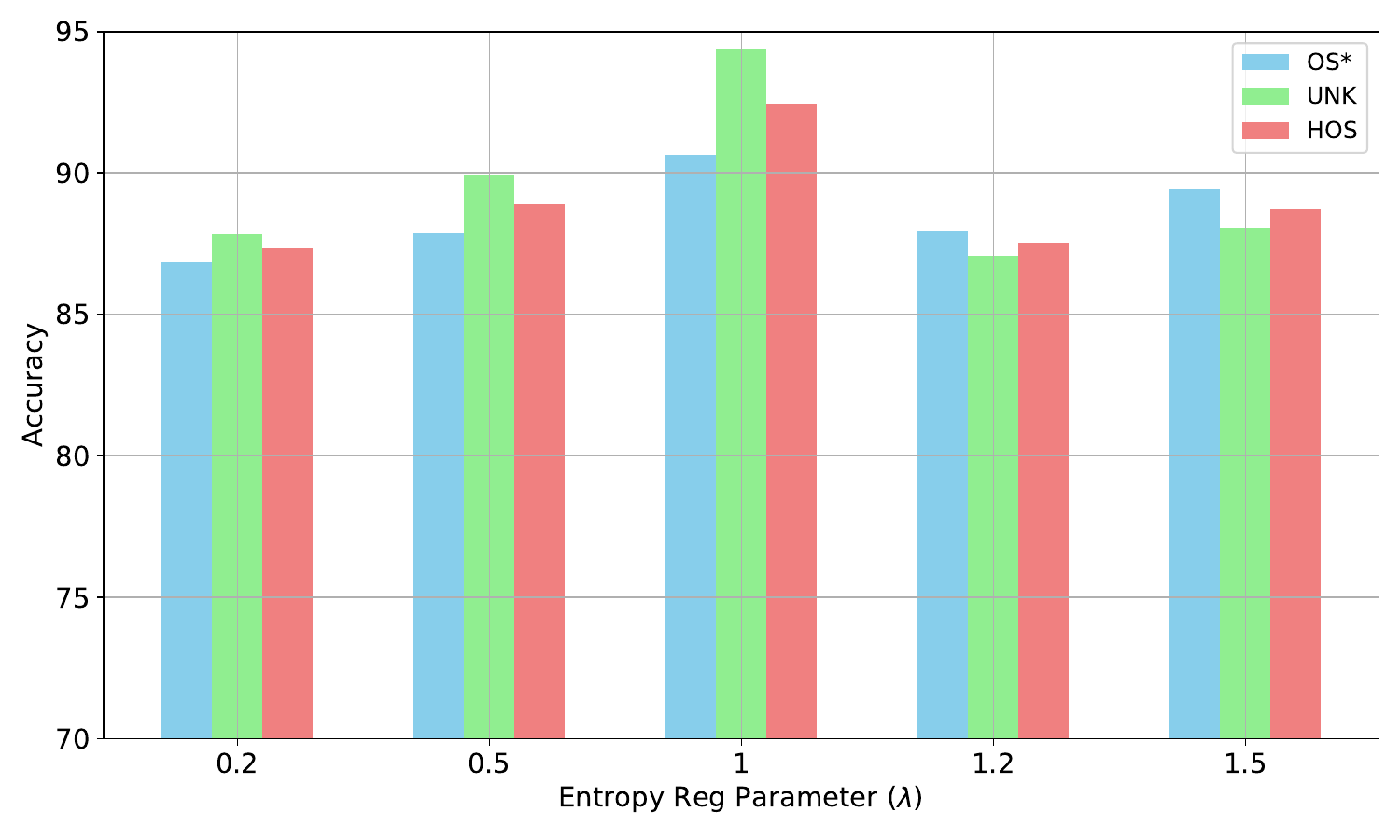}
    \vspace{-0.2cm}
\caption{Effect of varying the entropy regularization parameter $\lambda$ on the Office-31 dataset.}

    \label{fig:lambda_effect}
\end{figure}

\vspace{-0.2cm}
\noindent \textbf{Effect of entropy regularisation parameter (\( \lambda \)):}  Figure \ref{fig:lambda_effect} depicts the effect of varying the entropy regularization parameter (\( \lambda \)) on the model's metrics: OS*, UNK, and HOS. Optimal performance is achieved at \( \lambda = 1 \), suggesting that a balanced entropy regularization is crucial for enhancing model accuracy.

\begin{figure}[ht!]
    \centering
    \includegraphics[width=\textwidth]{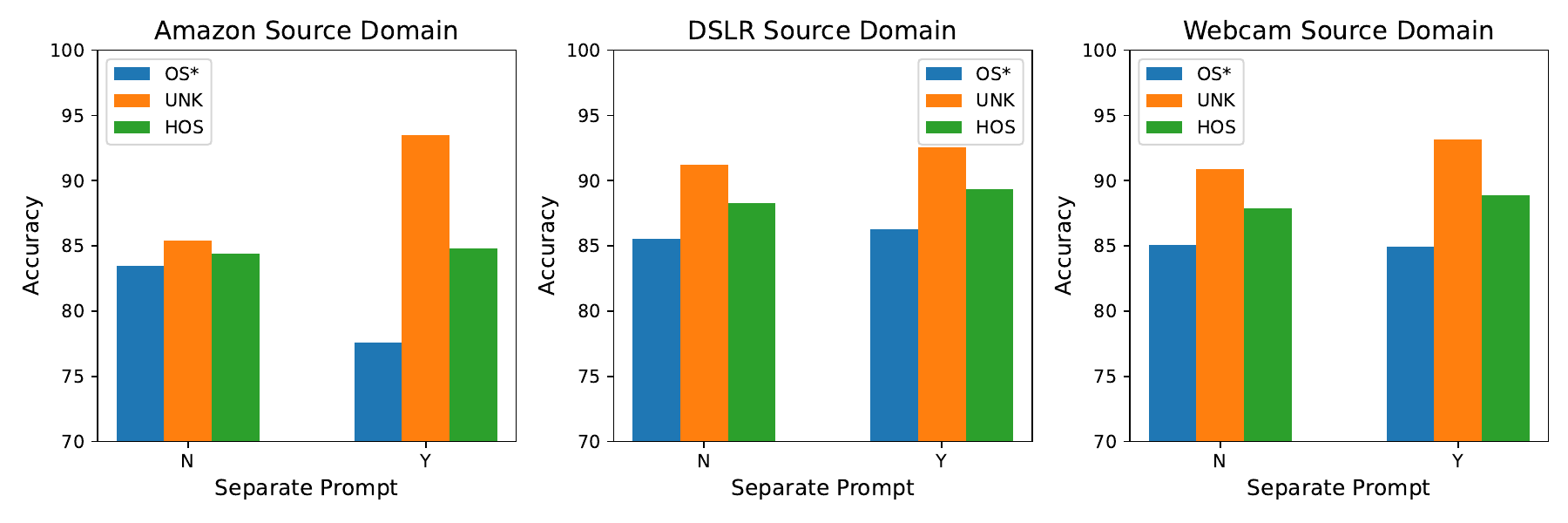}
    \vspace{-1cm}
\caption{Impact of having separate known prompts $P_{kwn}$ and unknown prompts $P_{unk}$. Here 'N' represents no separate prompt, and 'Y' represents that the separate prompts are used.}
    \label{fig:sep_prompts}
\end{figure}

\vspace{-0.2cm}
\noindent \textbf{Impact of having separate \(P_{kwn}\) and \(P_{unk}\):} Figure~\ref{fig:sep_prompts} provides a detailed analysis of the effects of using separate \(P_{kwn}\) and \(P_{unk}\) on the performance metrics: OS*, UNK, and HOS, across different source domains in the Office-31 dataset. The results demonstrate an increase in HOS scores (except on the Amazon domain) when separate prompts are implemented. A notable increase is observed in the Unknown accuracy, implying that separate prompts are able to handle the unknown classes well. 

\vspace{-0.8cm}
\section{Comprehensive Results}
\label{sec:results}
\vspace{-0.2cm}
\noindent\textbf{t-SNE visualization:} In Figure \ref{fig:tSNEs}, we visualize and compare the t-SNE embeddings generated from the text encoder of our proposed COSMo for both known and unknown classes with other methods on the Office-Home dataset on the proposed setting. COSMo is able to segregate the known and unknown classes better.
Tables \ref{table:results1}, \ref{table:results2}, and \ref{table:results3} depict the comprehensive results for the proposed setting on Office-31, Office-Home and Mini-DomainNet datasets, respectively. The results are obtained with both vision backbones: ResNet-50 \cite{resnet} and ViT-B/16 \cite{vit}, and all the metrics are reported (OS*, UNK and HOS). 

\begin{figure}[htbp]
    \centering
    \begin{tabular}{ccc}
        \includegraphics[width=0.3\linewidth, height=4cm]{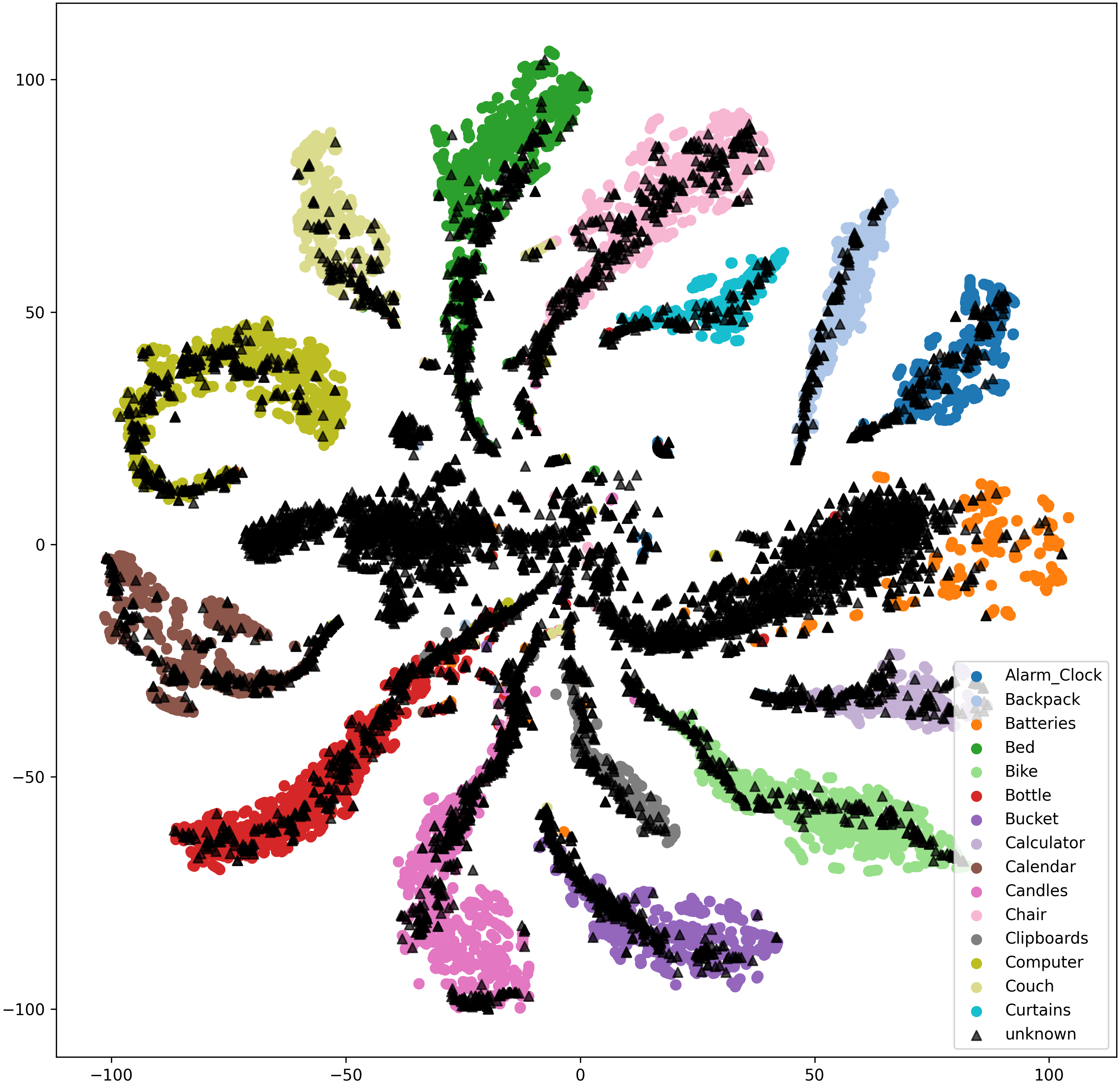} &
        \includegraphics[width=0.3\linewidth, height=4cm]{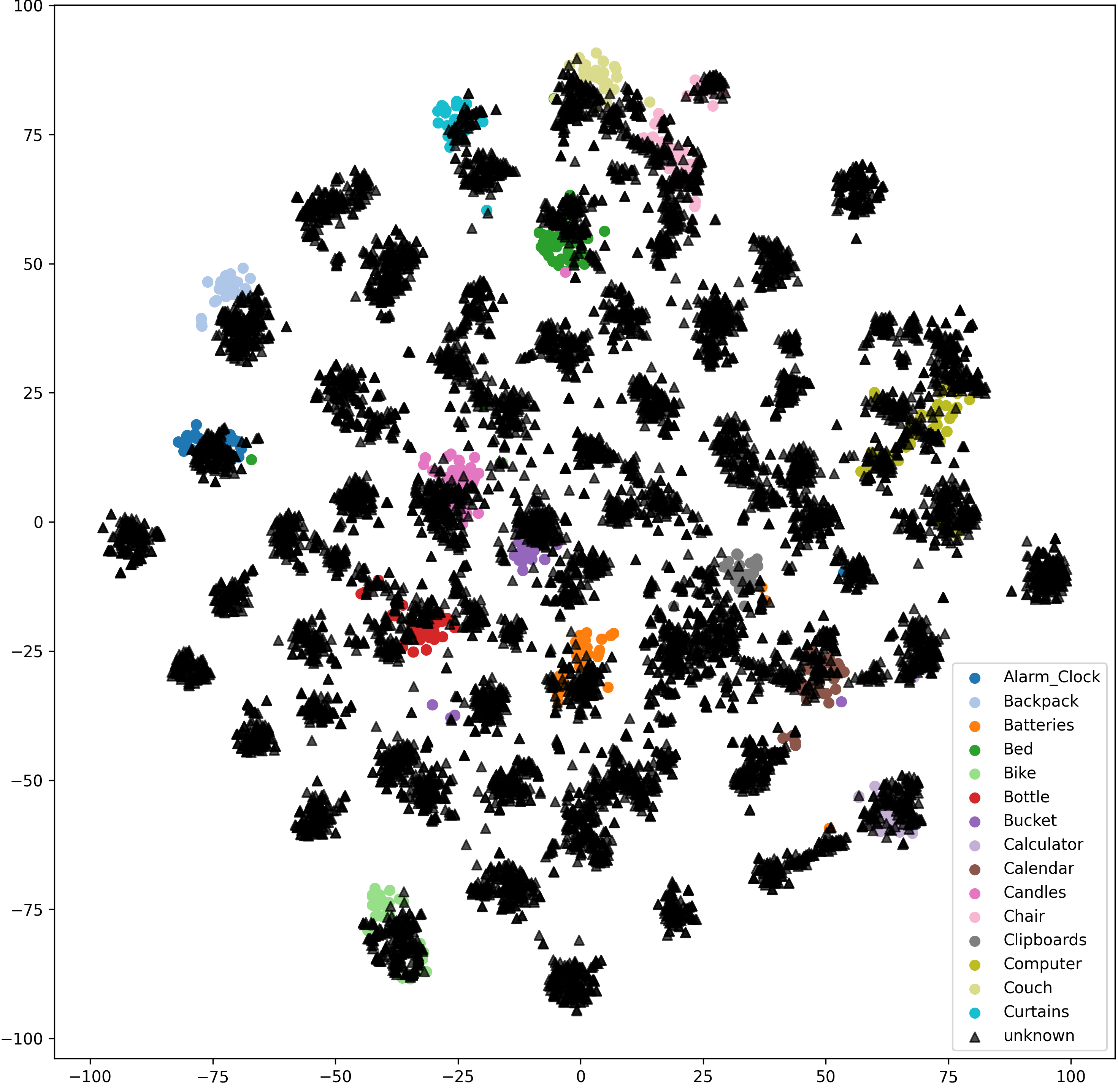} &
        {\includegraphics[width=0.3\linewidth, height=4cm]{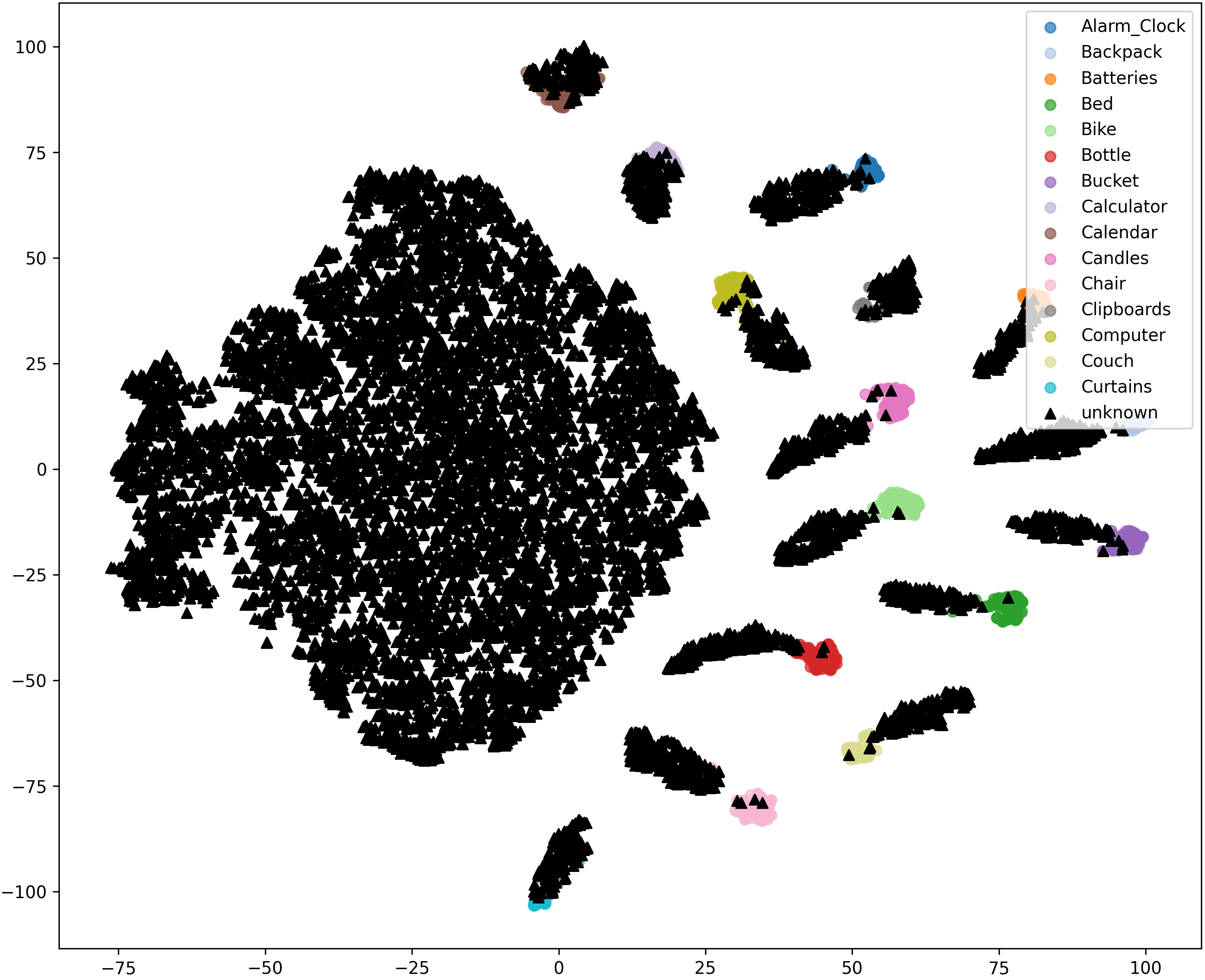}} \\
        
        (a) OSDA-BP & (b) DANCE & (c) COSMo (Ours) \\
    \end{tabular}
    \caption{
    t-SNE visualizations on the Office-Home Dataset with Amazon as the source domain. Coloured dots represent known classes in the source domain, while \textcolor{black}{black} triangles denote target domain samples. For COSMo, text embeddings are used, while features from the penultimate layer are used for the other models.}
    \vspace{-0.4cm}
    \label{fig:tSNEs}
\end{figure}

\noindent\textbf{Detailed results of Table 1 (Main paper):} Here, we discuss the detailed results of our proposed COSMo and compare them with the referred literature. 

\noindent Table \ref{table:results1} presents detailed results on the \textbf{Office-31} dataset. Our proposed COSMo outperforms other models across all three domains using the ViT-B/16 architecture for the OSMTDA task. With the ResNet 50 architecture, COSMo surpasses other models in two out of three domains. Additionally, COSMo achieves the highest HOS score on the Amazon domain and the lowest on the DSLR domain. 

\begin{table*}[ht]
\caption{Results on the Office-31 (10/21) dataset.  Best in \textbf{bold}, second best \underline{underlined}.}
\centering
\scalebox{1}{
\begin{tabular}{lccccccc}
\toprule
\multirow{2}{*}{\textbf{Method}} & \multirow{2}{*}{\textbf{Source Domain}}& \multicolumn{3}{c}{\textbf{RN50}}&\multicolumn{3}{c}{\textbf{ViT-B/16}}\\\cmidrule(lr){3-5}\cmidrule(lr){6-8}
& & \textbf{OS*} & \textbf{UNK} & \textbf{HOS} &\textbf{OS*} & \textbf{UNK} & \textbf{HOS} \\ \midrule
\multirow{3}{*}{CLIP \cite{clip}}  & Amazon & 92.53&	25.22&	39.64  & 95.24	&28.43&	43.79 \\
 &   DSLR & 89.34&	31.53&	46.61 &91.2&	25.58&	39.95 \\
 &   Webcam &  89.26&	30.72&	45.71  & 90.74	&25.33&	39.61 \\ \midrule

\multirow{3}{*}{OSDA-BP \cite{osda-bp}} & Amazon & 92.66& 25.11& 39.51& 90.24 & 78.43 & 83.92\\
 &   DSLR & 83.97 & 55.23 & 66.63 & 77.78 & 71.68 & 74.60 \\
 &   Webcam & 80.70 & 51.82 & 63.11 & 77.00 & 69.70 & 73.17 \\ \midrule
\multirow{3}{*}{DANCE \cite{dance}}  & Amazon & 96.02 & 64.05 & \underline{76.84} & 87.82 & 85.18 & \underline{86.48} \\
 &   DSLR & 78.12 & 83.00 & \underline{80.49}& 78.37 & 96.12 & \underline{86.34}  \\
 &   Webcam & 76.75 & 82.16 & \underline{79.36} & 80.84 & 95.98 & \underline{87.76} \\ \midrule
\multirow{3}{*}{AD-CLIP \cite{adclip}}  & Amazon &  93.26&	25.55&	40.11  & 100	&22.68&	36.97 \\
 &   DSLR & 92.3 &	22.76&	36.51 &79.73	&35.17	&48.81 \\
 &   Webcam &  90.46&	29.65&	44.66  & 92.07	&19.6&	32.32 \\ \midrule
\multirow{3}{*}{COSMo}  & Amazon & 74.58&	81.64&	\textbf{77.95} &90.64 &	94.36&	\textbf{92.46} \\
 &   DSLR & 82.93&	78.69&	\textbf{80.76} &87.05&	89.82&	\textbf{88.41} \\
 &   Webcam & 84.47&	76.94&	\textbf{80.53} & 87.75&	90.59&	\textbf{89.15} \\ \bottomrule
\end{tabular}}
\label{table:results1}
\end{table*}

\newpage
\noindent Similar to the results on the Office-31 dataset, Table \ref{table:results2} presents a detailed comparison of our proposed COSMo model with state-of-the-art methods on the \textbf{Office-Home} dataset. COSMo consistently outperforms nearly all other models across various domains, with the exception of the Art domain when using the ViT-B/16 architecture. The Art domain poses a greater challenge compared to the other domains. 

\begin{table*}[ht!]
\caption{Results on Office-Home (15/50) Dataset.  Best in \textbf{bold}, second best \underline{underlined}.}
\centering
\scalebox{1}{
\begin{tabular}{lccccccc}
\toprule
\multirow{2}{*}{\textbf{Method}} & \multirow{2}{*}{\textbf{Source Domain}}& \multicolumn{3}{c}{\textbf{RN50}}&\multicolumn{3}{c}{\textbf{ViT-B/16}}\\\cmidrule(lr){3-5}\cmidrule(lr){6-8}
& & \textbf{OS*} & \textbf{UNK} & \textbf{HOS} &\textbf{OS*} & \textbf{UNK} & \textbf{HOS} \\ \midrule

\multirow{4}{*}{CLIP \cite{clip}} & Art & 84.15&	48.00&	61.14 & 92.08&	46.06&	61.76 \\
 &   Clipart & 92.21&	45.28&	60.73 & 95.39	&45.01&	61.16\\
 &   Product & 81.44&	54.62&	65.38 & 90.64	&53.44&	67.24 \\
 &   Real World & 79.65&	51.09&	62.25 &90.16	&49.03&	63.52 \\ \midrule

\multirow{4}{*}{OSDA-BP \cite{osda-bp}} & Art & 75.36 &	29.88	& 42.8  &  42.44 &	75.7 &	54.39\\

 &   Clipart & 71.34 &	49.06	& 58.14 & 35.13	& 31.13 &	33.01\\
 &   Product & 67.71	& 41.82 &	51.71 & 54.77	& 57.23 &	55.97\\
 &   Real World & 68.14 &	43.39	& 53.02 & 48.32 &	69.33 & 56.95\\ \midrule
 
\multirow{4}{*}{DANCE \cite{dance}}  & Art & 67.91 &	81.08 &	\underline{73.91} & 79.63	 & 82.83 &	\textbf{81.2}\\
 &   Clipart & 65.14	& 84.49	& \underline{73.56} & 84.38 &	86.5 & \underline{85.42}\\
 &   Product & 58.8 &	85.73 &	\underline{69.76} & 72.57 &	85.68 &	\underline{78.58}\\
 &   Real World & 62.04	&81.82 &	\underline{70.57} & 77.5 &	81.13 &	\underline{79.28} \\ \midrule
\multirow{4}{*}{AD-CLIP \cite{adclip}}  & Art & 84.41&	47.43&	60.74 & 92.64 &	38.15&	54.04 \\
 &   Clipart & 92.3&	34.5	&50.23 & 94.02&	33.05&	48.91 \\
 &   Product & 82.11&	44.64	&57.84 &90.52&	34.23&	49.67\\
 &   Real World & 80.38&	42.36	&55.48 & 92.16 &	31.26&	46.69\\ \midrule
\multirow{4}{*}{COSMo} & Art &79.31&	74.74&	\textbf{76.96} &90.25&	73.4&	\underline{80.96}\\
 &   Clipart & 80.59&	81.99&	\textbf{81.28} &88.92	&84.79&	\textbf{86.8}\\
 &   Product &74.8&	70.42&	\textbf{72.54} & 80.78&	84.13&	\textbf{82.42}\\
 &   Real World &72.27&	75.47&	\textbf{73.83} &84.65&	79.44&	\textbf{81.97} \\
\bottomrule
\end{tabular}}
\label{table:results2}
\end{table*}

\newpage
\noindent The \textbf{Mini-DomainNet} dataset presents a significant challenge for domain adaptation due to its large number of unknown classes and the relatively high number of known and unknown samples. Despite these difficulties, our model achieves nearly 80\% HOS score, the highest among other models, across the four domains, as shown in Table \ref{table:results3}. Regardless of the architecture used, COSMo consistently attains the best HOS score across all four domains. Notably, we observe the highest HOS score on the sketch domain and the lowest on the real domain.

\begin{table}[ht]
\caption{Results on Mini-Domain Net (60/66) Dataset.  Best in \textbf{bold}, second best \underline{underlined}.}
\centering
\scalebox{1}{
\begin{tabular}{lccccccc}
\toprule
\multirow{2}{*}{\textbf{Method}} & \multirow{2}{*}{\textbf{Source Domain}}& \multicolumn{3}{c}{\textbf{RN50}}&\multicolumn{3}{c}{\textbf{ViT-B/16}}\\\cmidrule(lr){3-5}\cmidrule(lr){6-8}
& & \textbf{OS*} & \textbf{UNK} & \textbf{HOS} &\textbf{OS*} & \textbf{UNK} & \textbf{HOS} \\ \midrule
\multirow{6}{*}{CLIP \cite{clip}} & Clipart & 80.47&	64.59&	\underline{71.67} & 89.29&	65.81&	\underline{75.78} \\
 &   Painting & 82.15&	63.1&	\underline{71.38} & 90.97&	64.78&	75.67\\
 &   Real & 68.87&	68.19&	\underline{68.53} &84.74&	63.19&	72.39 \\
 &   Sketch & 81.07&	64.26&	\underline{71.69} & 89.67&	65.44&	\underline{75.66} \\
 \midrule
\multirow{6}{*}{OSDA-BP \cite{osda-bp}} & Clipart & 38.85&	71.14&	50.26&	37.43&	46.01&	41.27\\
 &   Painting & 38.58	&73.25	&50.54	&48.3&	57.28&	52.41\\
 &   Real & 25.73	&73.54	&38.12	&41.65&	59.08&	48.86\\
 &   Sketch & 38.76	&70.35	&49.98	&40.95&	35.09&	37.79\\
 \midrule
 \multirow{6}{*}{DANCE \cite{dance}} & Clipart & 26.46	&94.82&	41.37&	56.85&	91.96&	70.26\\
 &   Painting & 31.7	&94.29	&47.45	&76.07&	87.29&	\underline{81.29}\\
 &   Real & 14.39	&97.12	&25.06	&61.97&	89.85&	\underline{73.35}\\
 &   Sketch & 28.6	&96.27	&44.1&	60.46&	94.85&	73.84\\
 \midrule
  \multirow{6}{*}{AD-CLIP \cite{adclip}} & Clipart & 83.97&	41.93&	55.93 & 91.87&	37.4&	53.16 \\
 &   Painting & 85.38&	44.18&	58.23 & 92.98	&41.01&	56.92 \\
 &   Real & 73.5&	40.36&	52.11 &86.12&	32.9&	47.61 \\
 &   Sketch & 83.1&	43.67&	57.25 & 91.94&	38.1&	53.88 \\
 \midrule
 \multirow{6}{*}{COSMo} & Clipart & 74.38&	76.62&	\textbf{75.48} &83.08&	79.1&	\textbf{81.05} \\
 &   Painting &80.37&	72.69&	\textbf{76.34} & 86.58&	81.85&	\textbf{84.15} \\
 &   Real & 63.36&	77.06&	\textbf{69.54} &79.33&	79.03&	\textbf{79.18} \\
 &   Sketch & 72.75&	80.72&	\textbf{76.52} &81.08&	84.77&	\textbf{82.89} \\
 
 \bottomrule
\end{tabular}}
\label{table:results3}
\end{table}

\begin{table}[ht!]
\centering
\caption{Table of Mathematical Terms and Notations}
\scalebox{1}{
\begin{tabular}{ll}
\toprule
\textbf{Notation} & \textbf{Description} \\ 
\midrule
\( (X_s, Y_s) \in S \) & Source domain data and labels \\ 
\( X_t \) & Unlabeled data from all target domains combined \\ 
\( q \) & Number of target domains \\ 
\( C_s \) and \( C_t \) & Classes in the source and target domains \\ 
\( C_k \) & Known classes from the source domain, \( C_k = C_s \) \\ 
\( C_u \) & Unknown classes in the target domain, \( C_u = C_t \setminus C_s \) \\ 
\( D_s \) and \( D_t \)& Mini-batch from labeled source and unlabelled target domains \\ 
\( \mathcal{F}_v \) and \( \mathcal{F}_t \) & Pre-trained image and text encoder \\ 
\( B_\theta(\cdot) \) & Domain-specific bias network \\ 
\( \beta \) & Domain bias context token, \( \beta = B_\theta(v) \) \\ 
\( P_k^c \) & Known class-based prompt for class \(c\) \\ 
\( P_{k,\ bias}^c \) & Biased known class prompt \\ 
\( P_{kwn} \) & Cumulative prompt for all known classes \\ 
\( P_u \) & Unknown class-based prompt \\ 
\( P_{unk} \) & Biased unknown class prompt \\ 
\( \tau \) & Text features encoded by the text encoder \\ 
\( \lambda \) & Hyperparameter controlling entropy regularization strength \\ 
\( \kappa_{lower} \) & Lower threshold for confidence in known classes \\ 
\( \kappa_{upper} \) & Upper threshold for confidence in unknown classes \\ 
\( m \) & Length of the context prompt \\ 
\bottomrule
\end{tabular}}
\label{tab:notations}
\end{table}

\end{document}